%% file: 285.tex
\documentclass{llncs}
\usepackage[T1]{fontenc}
\usepackage[utf8]{inputenc}
\usepackage{lmodern}
\usepackage{algorithm}
\usepackage{algorithmicx}
\usepackage[noend]{algpseudocode}
\usepackage{amssymb}
\usepackage{graphicx}
\usepackage{tikz}
\usepackage{mathtools}
\usetikzlibrary{arrows,shapes,shadows,positioning,fadings,decorations,decorations.pathmorphing}
\tikzstyle{block}=[draw opacity=0.7,line width=1.4cm]
\newcommand{\subtreerootedat}[1]{\raisebox{-0.1cm}{$\stackrel{#1}{\triangle}$}}

\title{Extending local features with contextual information in graph kernels}
\author{Nicol\`o Navarin \and Alessandro Sperduti \and Riccardo Tesselli} 
\institute{Department of Mathematics, University of Padova, Italy \\\email{\{nnavarin,sperduti\}@math.unipd.it}\\ \email{rtessell@studenti.math.unipd.it}}
\begin{document}

\maketitle

\begin{abstract}
Graph kernels are usually defined in terms of simpler kernels over local substructures of the original graphs.
Different kernels consider different types of substructures.
However, in some cases they have similar predictive performances, probably because the substructures can be interpreted as approximations of the subgraphs they induce. 
In this paper, we propose to associate to each feature a piece of information about the context in which the feature appears in the graph.
A substructure appearing in two different graphs will match only if it appears with the same context in both graphs.
We propose a kernel based on this idea that considers trees as substructures, and where the contexts are features too.
The kernel is inspired from the framework in~\cite{Dasan2012}, even if it is not part of it.
We give an efficient algorithm for computing the kernel and show promising results on real-world graph classification datasets.
\end{abstract}
\input{sec_introduction}
\input{sec_graphkernels}
\input{section_contexts}
\input{sec_implementation_new}
\input{experiments}
\input{sec_conclusions}
\subsubsection{Acknowledgments.} This work was supported by the University of Padova under the strategic project BIOINFOGEN. 
\bibliographystyle{abbrv}
\bibliography{Mendeley}
\end{document}

%% file: sec_introduction.tex
\section{Introduction}
In many application domains data can be naturally represented in a structured form, e.g. in Chemoinformatics~\cite{Aggarwal2010} or in natural language processing~\cite{Collins2001}.
For this reason, in the last few years an interest in machine learning techniques applicable to data represented in structured (non-vectorial) form arose \cite{Vishwanathan2010,MartinoS10}.
When dealing with machine learning for graph-structured data, kernel methods are one of the most popular approaches to follow. 
It just suffices to use a kernel for graphs together with any kernelized learning algorithm (e.g. SVM, SVR, KPCA, \ldots) and the user has a powerful, ready-to-use learning algorithm with strong theoretical bounds on its generalization performance.
The predictive performance of the resulting learning procedure strongly depends on the particular kernel choice.
The design of efficient graph kernels is not a trivial task, because several graph operations (e.g. the graph isomorphism)
are not efficiently computable.
The idea is to design kernels that are the most expressive as possible, in order to have a small information loss.
Several alternatives have been proposed in literature. However it is difficult to state a priori which kernel will perform better in a specific task, because most of the existing kernels consider different approximations of the same local structures.
In this paper, we propose a method to enrich the feature space of a kernel with contextual information, i.e. we attach to a feature a piece of information about the topology of the graph in which that feature appeared.
We apply this idea to the ODD kernel~\cite{Dasan2012}, and we define as the context of a feature another feature from the same kernel. We give an efficient algorithm for the kernel computation, and experimentally evaluate our proposal on five real-world datasets.

%% file: sec_graphkernels.tex
\section{Definitions and notation}
Let $G=(V_G,E_G,L_G)$ be a graph, where $V_G$ is the set of vertices (or nodes), $E_G\subseteq\{(v_i,v_j) | v_i,v_j \in V_G\}$ is the set of edges and $L_G:V_G \rightarrow \Sigma$ is a labeling function mapping each vertex to an element in a fixed alphabet $\Sigma$.\\
A graph is undirected if $(i,j) \in E_G \implies (j,i)\in E_G$, otherwise it is directed.
A walk $w(u,v)$ in a graph is a sequence of nodes $v_1, \ldots,v_n$ s.t.
 $(v_i,v_{i+1}) \in E_G$ and $v_1=u, v_n=v$.
The length of a walk $|w(u,v)|$ is defined as the number of edges in such walk.
A cycle is a walk where $v_1=v_n$. A graph is acyclic if it does not contain cycles.
A DAG is a directed acyclic graph.
A path is a walk with no repeated nodes, i.e. where $\forall_{i=1}^n \forall_{j=1}^n, i \neq j \implies v_i \neq v_j$.
A shortest path between two vertices $sp(u,v) \in V_G$ is a path with the minimum length that starts from $u$ and ends in $v$. Note that the shortest paths are not unique, but their length $|sp(u,v)|$ is.
$n\_sp(u,v)$ is a function returning the number of such shortest paths. 
\noindent A rooted DAG $D$ is a DAG in which one vertex $r$ has been designated as the root. The root have no incoming edges, i.e. $\not\exists u \in E_D, (u,r) \in E_D$. The function $r(D)$ returns the root of a rooted DAG.\\
A (rooted) tree is a rooted DAG where for each node there exists exactly one path connecting the root node to it. 
The children $children(v)$ of a node $v \in V_T$ in a tree are all the nodes $u \in V_T$ s.t. 
$(v,u) \in E_T$.
The number of children, or out-degree, of a vertex $v$ is $\rho(v)$.
Similarly we can say that $v$ is a parent
 of $u$.
$ch_i(v,G)$ is the function retuning the i-th child of $v \in V_G$ (according to a particular order).\\
A proper subtree rooted at $u\in V_T$ of a tree $T$ is the subtree that comprehends $u$ and all its descendants. We will refer to it as $\subtreerootedat{u} \in T$.
%
We define $T_j(v,G)$ as a function returning the tree-visit of a graph $G$, rooted at $v$ and limited at height $j$. Note that this tree-visit is the shortest-path tree between $v$ and any $u \in V_G$ s.t. $|sp(u,v)|\leq j$.
Moreover, we denote with $T(v,G)$ the tree-visit at the maximum possible height, i.e.  $T_\infty(v,G)=T_{diam(G)}(v,G)$ where $diam(G)$ is the diameter of a graph, i.e. the length of the longest shortest path between two vertices.\\
A DAG-visit of a graph $G$, $DAG_j(v,G)$, is defined as the DAG of the shortest paths of length up to $j$. The main difference between $DAG(v,G)$ and $T(v,G)$ is that the number of nodes in the former is bounded by $|V_G|$ while in the latter it is not.
We assume the nodes in $T_j(v,G)$ or $DAG_j(v,G)$ to be ordered according to the lexicographic order between the node labels (in case two nodes have the same label, the ordering is recursively induced from the children). Such an ordering has been proven to be well-defined in~\cite{Dasan2012} for DAGs. Since trees are a special case of DAGs, the ordering relation is well-defined on trees as well.
For ease of notation, when clear from the context, the link to the graph $G$ will be omitted from the above-mentioned functions.
\section{Graph Kernels}
%
Most of the existing graph kernels are members of the $R$-convolution kernels framework~\cite{Haussler99convolutionkernels}.
The idea of this framework is to decompose the original structure into a set of simpler structures, where a (efficient) kernel is already defined.
For example, the all-subgraphs graph kernel~\cite{Gartner2003a} has a feature associated to each possible graph. However, this kernel also happens to be NP-complete.
An approach to reduce the computational complexity of the resulting kernel is to restrict the set of considered substructures of the graph. Different substructures raise different kernels. For example, in literature kernels based on random walks~\cite{Vishwanathan:uq}, shortest paths~\cite{Kriegel05shortestpath}, subtree-patterns~\cite{ShervashidzeSLMB11}, subtrees~\cite{Dasan2012} or pairs of small rooted subgraphs \cite{Costa2010} have been proposed.
The main drawback of these kernels is that they consider only local substructures of the original graphs, whose size is bounded to some limit due to computational complexity.
For this reason, in some cases they have similar predictive performances \cite{DaSanMartino2015}, probably because the different substructures can be interpreted as different, but still similar, approximations of small subgraphs of the original graph.
Enlarging the substructures to let the kernel consider a larger amount of information will increase the computational burden. We recall that the main challenge while designing graph kernels is the trade-off between the efficiency and the expressive power of the kernel.\\
Among the available graph kernels, the NSPDK~\cite{Costa2010} is the most related to the proposed kernel. Specifically, in the RKHS of NSPDK, every feature represents a couple of small rooted subgraphs $S_1$ and $S_2$ of a certain diameter (radius) $r$, at a certain distance, i.e. where $|sp(r(S_1),r(S_2))|=d$. In a sense, $S_1$ can be seen as a context for $S_2$ and vice versa.\\
%
%
%
%
%
Let us define a set of Ordered Decomposition DAGs of a graph $G$ limited to the maximum (user-specified) depth $h$ as $ODD_G=\{DAG_h(v,G) | v \in V_G\}$, where we recall that the nodes in each DAG are ordered according to a recursive relation looking at the labels of a node and all its descendants.
The $ODDK$ kernel~\cite{Dasan2012} is defined as:
\[
ODDK(G_1,G_2)=\sum_{\substack{OD_1 \in ODD_{G_1}\\OD_2 \in ODD_{G_2} }} \sum_{j=1}^h \sum_{l=1}^h\sum_{\substack{v_1 \in V_{OD_1}\\v_2 \in V_{OD_2}}} C_{ST}(r(T_j(v_1)),r(T_l(v_2)))
\]
where $C_{ST}()$ is a function that defines the subtree kernel, i.e. a kernel that counts the number of shared proper subtrees between two trees.
This kernel allows to obtain an explicit feature space representation $\phi$~\cite{DaSanMartino2012}.
Let us define a total ordering between all the possible labeled trees that appear from the kernel application on the dataset.
Then each feature $\phi_i(G)$ represents the frequency of the $i$-th tree in the RKHS of the $ODD$ kernel.

%% file: section_contexts.tex
\section{Adding Contexts to Graph Kernels\label{sec:ODDWCK}} 
The graph kernels described in the previous section extracts local patterns of the graph as features, i.e. the feature itself does not bring any information regarding where it has appeared within the graph. The idea we propose in order to increase the expressiveness of a kernel, while preserving efficiency, is to enrich the local features (e.g. the features extracted by the $ODD$ kernel) with their contextual information. The contextual information we are interested in is a description of the topology of the graph around the extracted feature. Thus, a substructure that appears in two different graphs will match if and only if it appears within the same context in both graphs.
Considering contextual information, we obtain kernels that are more sparse.
In some cases, the resulting kernel may be more discriminative with respect to the original one.
However, in other cases it may be too much sparse to obtain good performance. In the latter case, it can be beneficial to add the contribution of the new kernel to the original one. In our experiments, we will implement both these variants. Note that, with our proposed approach, the computation of the contributions of the contextualized kernel and of the original kernel can be performed efficiently at the same time.\\
Fixed a feature of the original graph kernel, we want the following property to hold:
$$
\sum_{c\in Contexts(f)} \phi_{f\circ c}(G) = \phi_f(G),
$$
where $\phi_f(G)$ is the frequency of a feature $f$ in the RKHS of the original kernel, and $\phi_{f \circ c}$ is the frequency of $f$ appearing within the context $c$.
From the formula it is clear that for each feature we need to consider also the empty context($\varnothing$-context), i.e. the situation in which a feature does not appear in any particular context e.g. because it has reached the maximum allowed dimension and we have no information about its context in the original graph.\\
%
%
%
In the remaining of this section, we will introduce our proposed kernel instantiating the context idea to the $ODD$ kernel.
As a feature represents a substructure, in the same way we can represent a context for a feature as a substructure of the graph, that incorporates the feature. Therefore, contexts and features can share the same representation and so it is possible that a context for a given feature can be a feature itself.
To compute the contextualized features we only need to combine a feature with other features representing the context in which the first feature appears in the graph.\\
The first important difference between the proposed Tree Context Kernel (TCK) and ODDK is that, for technical reasons, the former is defined over tree-visits while the latter over DAG-visits.
Note that the nodes of a tree-visit $T(v,G)$ of a graph $G$ can grow exponentially in its size, while if we consider a DAG-visit $DAG(v,G)$, each node in the original graph can appear at most once, thus limiting the size of the resulting structure to at most $|V_G|$ nodes.
However, in the next section we will provide an efficient implementation that does not need to store in memory the tree-visits, but only the DAG-visits.
The Tree Context Kernel can be defined as:
\begin{eqnarray}
TCK(G_1,G_2) & = & \sum_{\substack{v_1 \in V_{G_1}\\v_2 \in V_{G_2} }}\sum_{i=1}^h \sum_{j=1}^h \nonumber\\ 
&  & [ \delta(T_i(v_1),T_j(v_2)) + \!\!\!\!\!\!  \sum_{\substack{\subtreerootedat{u_1} \in T_i(v_1)\\\subtreerootedat{u_2} \in T_j(v_2)}}\!\!\!\!\delta(\subtreerootedat{u_1}, \subtreerootedat{u_2})  \sum_{l=1}^{\rho(u_1)}C_{ST}(ch_l(u_1), ch_l(u_2))]\nonumber
\end{eqnarray}
\noindent where we recall that:
\[
C_{ST}(v_1,v_2)=\\
\begin{cases}
\lambda \cdot K_L(v_1,v_2) & \!\!\!\!\!\!\!\!\!\!\!\!\!\!\!\!\!\!\textrm{if }v_1 \textrm{ and } v_2 \textrm{ are leaves}\\
\lambda \cdot K_L(v_1,v_2) \prod_{j=1}^{\rho(v_1)}C_{ST}(ch_j(v_1),ch_j(v_2)) & \textrm{if } \rho(v_1)=\rho(v_2) \\
0 &\text{otherwise} \\
\end{cases}
\]
and $\delta$ is the Kronecker's delta function.
We recall that $C_{ST}(v_1,v_2), v_1 \in T_1, v_2 \in T_2$ is a function that counts the common proper subtrees of two trees.
The function depends on $T_1$ and $T_2$. We decided to follow the original definition of~\cite{Collins:2001fk} omitting that dependency for ease of notation.
\\
The kernel is positive semidefinite because it is a composition of positive semidefinite kernels, defined over the ordered tree visits $T_i(v,G)=T(DAG_i(v,G))$ that are well defined as shown in~\cite{Dasan2012}.\\
Intuitively, this kernel matches two subtree features $\subtreerootedat{u_1} \in T_i(v_1,G_1), 0 \leq i \leq h$ and $\subtreerootedat{u_2} \in T_j(v_2,G_2), 0 \leq j \leq h$ in one of the following cases:
\begin{itemize}
  \item both $v_1$ and $v_2$ are the root nodes of the tree visit, i.e. $u_1=v_1$ and $u_2=v_2$;
  \item $u_1$ and $u_2$ occur within the same context in both trees, i.e. their parents generate the same proper subtree.
\end{itemize}

%% file: sec_implementation_new.tex
\section{Efficient Implementation}
Algorithm \ref{alg:decomposeexpl} shows the pseudocode to decompose a graph $G$ into its explicit (sparse) feature vector $\phi$. 
We will denote with $f$ the map that stores the keys of the local subtree features, i.e. $f_{u,d}, u \in V_G, d \in \{0,\ldots,h\}$ is the key of the subtree rooted in $u$ of height $d$. Similarly, $size$ is a map that stores the size of each feature, i.e. $size_{u,d}$ is the number of nodes that compose the feature $f_{u,d}$. Let $\kappa$ be a perfect hash function from strings to integers.
Such a function can be implemented with an incrementally-built hashmap that associates an unique id to each string. Alternatively, a normal hashing function can be used if we tolerate some clashes.  We define reserved special symbols ``$ \lceil$'', ``$\rfloor$'', ``\#'' and ``$\circ$'' that do not have to appear in the labels of the graphs and they are needed to encode subtree features into strings.
\begin{algorithm}[t]
\caption{An algorithm for computing the explicit feature space representation of a graph $G$ according to the kernel $TCK_{ST}$ with maximum (user-specified) height $h$ and weight factor $\lambda$}\label{alg:decomposeexpl}
\begin{algorithmic}[1]
\State $\phi=[0,\dots,0]$ \Comment{Explicit feature space represented as sparse vector}
\ForAll{$v \in V_G$}
\State $D \gets DAG_{h}(v,G)$ 
\State $f=\{\}$ \Comment{dictionary that stores the features related to a node $u$ and height $d$}
\State $size=\{\}$ \Comment{dictionary that stores the size of each feature}
\ForAll{$u \in$ \Call{reverseTopologicalOrder}{$D$}}
\For {$d\gets 0,\ldots,diam(D)-|sp(v,u)|$}
\If{$d=0$}
\State $f_{u,0}\gets \kappa(L(u))$
\State $size_{u,0}\gets 1$ 
\Else
\State $(S_1,\dots,S_{\rho(u)})\gets$ \Call{sort}{$f_{ch_1(u),d-1},f_{ch_2(u),d-1},\cdots,f_{ch_{\rho(u)}(u),d-1}$}
\State $f_{u,d}\gets \kappa(L(u) \lceil S_1 \#  S_2 \#\cdots \# S_{\rho(u)}\rfloor)$
\State $size_{u,d}\gets 1 + \sum_{i=1}^{\rho(u)}size_{ch_i(u),d-1}$
\ForAll{$ch \in children(u)$} 
\State $\phi_{f_{ch,d-1}\circ f_{u,d}}\gets \phi_{f_{ch,d-1}\circ f_{u,d}}+ n\_sp(v,u)\cdot \lambda^{\frac{size_{ch,d-1}}{2}}$
\EndFor
\EndIf
\If{$u=v$}
\State $\phi_{f_{u,d}\circ \varnothing}\gets \phi_{f_{u,d}\circ \varnothing} + \lambda^{\frac{size_{u,d}}{2}}$ 
\EndIf
\EndFor
\EndFor
\EndFor
\State \Return $\phi$
\end{algorithmic}
\end{algorithm}
 In the following, we will discuss the most sensitive steps of the algorithm. In line 6 the nodes of the DAG-visit are traversed in a reverse topological order, ensuring that every node will be processed before its parent.
In line 7, for each node $u$ of the current DAG-visit $D$, we consider all the 
heights for the feature generation. Note that when $d=0$, $f_{u,0}$ is a feature (proper subtree) of the tree $T_{|sp(v,u)|}(v)$ and when $d=diam(D)-|sp(v,u)|$, $f_{u,d}$ is a feature of $T_{diam(D)}(v)$, where $diam(D)\leq h$.
Notice that if $D$ is unbalanced and we are considering a node $u$ whose $|sp(v,u)|$ is not maximum, then we are considering many times the feature associated to $u$ at its maximum height.
In lines 12-14, the local feature related to the current node and height is generated. The hashed feature values of the children of the current node at height $d-1$ are sorted, generating a feature of height $d$  and inducing an order on the children of every node that is the lexicographic order over the hash values of the corresponding features. This step allows us not to define any particular ordering on the nodes of $D$. Then the extracted feature is encoded and finally it is hashed.
Lines 15-16 generate the contextualized features and increment their frequency in $\phi$ according to
a weight term multiplied by $n\_sp(v,u)$. This multiplication allows us to compute the statistics related to the tree-visit while working on the smaller (in terms of number of nodes) corresponding DAG-visit. Notice that $n\_sp(v,u)$ is efficiently computed during the creation of $DAG_{h}(v,G)$ in a top-down fashion without any additional cost. 
Finally, lines 17-18 increment the value corresponding to the feature with empty context $\phi_{f_{u,d}\circ \varnothing}$. This implementation returns the explicit sparse feature vector $\phi$, therefore in order to compute the kernel function between two graphs is sufficient to compute the dot product between the two feature vectors.

%% file: experiments.tex
\section{Experimental results}
We measured the predictive performance of TCK and other state-of-the-art kernels on the following real-world datasets: AIDS, CAS, CPDB, GDD and NCI1. Each dataset represents a binary classification problem
and is composed by labeled graphs with no self-loops. The AIDS, CAS, CPDB and NCI1 datasets are collections of chemical compounds represented as graphs, with nodes labeled according to the atom type and edges that represent the bonds. The GDD dataset is composed by proteins represented as graphs, where the nodes 
represent amino acids and two nodes in a graph are connected by an edge if they are less than $6$\,\AA\, apart. The largest datasets are CAS and NCI1 with more than 4000 graphs, and the smallest is CPDB with 684 instances.\\
Since we cannot know in advance whether the sparsity is beneficial for a particular task, we choose to test two versions of the proposed kernel. 
The first version ($TCK$) considers only contextualized features, while the second version ($TCK+ODDK$) combines $TCK$ with the base (non-contextualized) kernel, $ODDK$ in our case.
Note that $TCK+ODDK$ can be computed with a slight modification of Algorithm~\ref{alg:decomposeexpl}, thus the computational complexities of the two versions of the proposed kernels are the same.
\noindent We compare the proposed kernels with the NSPDK kernel~\cite{Costa2010}, the Fast Subtree Kernel (FS)~\cite{ShervashidzeSLMB11}, and the original version of the ODDK based on the subtree kernel~\cite{Dasan2012}. To assess the predictive performances of the different kernels, we used a \textit{nested} 10-fold cross validation: within each of the 10 folds, another 10-fold cross validation is performed over the corresponding training set in order to select the best parameters for the current fold. Thus, the parameters are optimized on the training dataset only.
The whole process has been repeated 10 times using different random data splits.
The parameter space for both versions of $TCK$ and $ODDK$ was restricted to the following values: $h=\{1,2,\dots,10\}$ and 
$\lambda=\{0.1,0.5,0.8,0.9,\ldots,1.5,1.8\}$.
The parameter $h$ of the FS kernel were restricted to $h=\{1,2,\dots,10\}$ and for the NSPDK the values $h=\{1,2,\dots,8\}$ and $d=\{1,2,\dots,7\}$ were considered.
The SVM solver had the $C$ parameter ranging in $C=\{10^{-4},10^{-3},\ldots,10^{3}\}$.
\begin{table}[t]
    \caption{Accuracy results of the proposed kernels and the considered baselines, in nested 10-fold cross validation.}\label{Tab:nested}
\centering
    \begin{tabular}{ | l | c | c | c | c | c |}
    \hline
        \textit{Kernel/dataset} & {CAS} & {GDD} & {NCI1} & {AIDS} & {CPDB}  \\ \hline
    $NSPDK$ & $83.6_{\pm0.34}$ & 74.09$_{\pm0.91}$ &  $83.46_{\pm0.46}$ & $82.71_{\pm0.66}$ & $76.99_{\pm1.15}$  \\
    $WL$ & $83.33_{\pm0.37}$ & $75.29_{\pm1.33}$ &  $84.41_{\pm0.49}$ & $82.02_{\pm0.4}$ & $76.36_{\pm1.4}$  \\
    $ODDK$ & $83.53_{\pm0.21}$ & $76.99_{\pm0.36}$ &  $85.31_{\pm0.26}$ & \textbf{82.99}$_{\pm0.50}$ & $78.44_{\pm0.76}$ \\ \hline
    $TCK$ & $83.53_{\pm0.32}$ & \textbf{79.35}$_{\pm0.45}$ & \textbf{85.78}$_{\pm0.22}$ & $82.88_{\pm0.39}$ & $76.96_{\pm0.96}$ \\
    $TCK+ODDK$ & \textbf{83.94}$_{\pm0.26}$ & $78.03_{\pm0.56}$ & $85.48_{\pm0.182}$ & $82.97_{\pm0.5}$ & \textbf{78.89}$_{\pm0.98}$ \\
 \hline
    \end{tabular}

\end{table}
\begin{figure}[t]
\centering
  \includegraphics[width=0.72\textwidth]{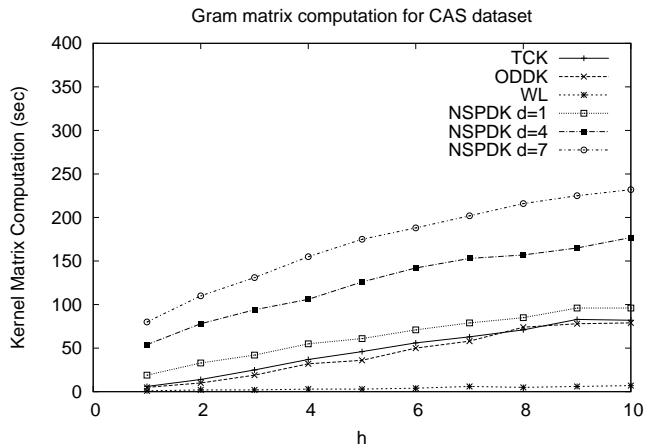}
  \caption{Copmuptational time (in seconds) required for the Gram matrix computation of the considered kernel, with different parameters.\label{fig:times}}
\end{figure}
Table~\ref{Tab:nested} reports the averaged accuracy results of our experiments with the corresponding standard deviations.
At a first glance, it is clear that in almost all the considered datasets, one of the two proposed kernels is the better performing among all the considered kernels, with the only exception of the AIDS dataset.
Looking at the results in more detail, in two datasets (GDD, NCI1) both versions of $TCK$ perform better than the others. If we consider the CAS dataset, the performance of the worst of the proposed kernels is comparable with the better kernel among the baselines (NSPDK). In the CPDB dataset the worst of the proposed kernels
 is worse than the best kernel among the baselines (ODDK), but it 
 is still competitive, such as in AIDS dataset, where the proposed kernels are very close to the best one.
\noindent Let us finally anlyze the computational requirements of our proposed kernel. Figure~\ref{fig:times} reports the computational times required for the Gram matrix computation of the kernels considered in this section on the CAS dataset.
The execution times of the proposed kernel are very close to the ones of the original kernel. The situation is similar for other datasets, and thus the corresponding plots are omitted.
%
The results presented in this section suggest that the introduction of contextualized features is a promising approach, and that in principle also other kernels can benefit from such an extension.

%% file: sec_conclusions.tex
\section{Conclusions and future work}
In this paper, we proposed a technique to incorporate context information in the kernels that allow for an explicit feature space representation.
In particular, we defined a relationship between the explicit features where one feature can be considered as the context of another one.
We applied our idea to the $ODDK$ kernel, and slightly modified the kernel definition in order to provide an efficient algorithm for the computation of the contextualized kernel.
We evaluated the predictive performance of the resulting kernel (in two variants) over five real-world datasets, and the proposed approach shows promising results.
As future works, we plan to apply the contextualization idea to other state-of-the art graph kernels, as well as to kernels for other discrete structures.